\documentclass[conf]{new-aiaa}
\usepackage[utf8]{inputenc}

\usepackage{xcolor}
\usepackage{lineno,hyperref}
\modulolinenumbers[5]

\newcommand{\eg}{{\em e.g.}}
\newcommand{\ie}{{\em i.e.}}

\makeatletter
\newcommand*{\rom}[1]{\expandafter\@slowromancap\romannumeral #1@}
\makeatother

\usepackage{graphicx}
\usepackage{amsfonts}
\usepackage{amsmath}

\usepackage[version=4]{mhchem}
\usepackage{siunitx}
\usepackage{longtable,tabularx}

\usepackage{bm}
\usepackage{multirow}
\usepackage{multicol}
\usepackage[colorinlistoftodos]{todonotes}
\usepackage{tabularx} 
\usepackage{gensymb}

\hypersetup{
    colorlinks,
    linkcolor={red!50!black},
    citecolor={blue!50!black},
    urlcolor={blue!80!black}
}

\title{Deep Generative Model for Efficient 3D Airfoil Parameterization and Generation}

\author{Wei Chen\footnote{Research Scientist; chen.wei@siemens.com.} and Arun Ramamurthy\footnote{Research Scientist.}}
\affil{Siemens Corporate Technology, Princeton, New Jersey, 08540}

\begin{document}

\maketitle

\begin{abstract}
In aerodynamic shape optimization, the convergence and computational cost are greatly affected by the representation capacity and compactness of the design space. Previous research has demonstrated that using a deep generative model to parameterize two-dimensional (2D) airfoils achieves high representation capacity/compactness, which significantly benefits shape optimization. In this paper, we propose a deep generative model, Free-Form Deformation Generative Adversarial Networks (FFD-GAN), that provides an efficient parameterization for three-dimensional (3D) aerodynamic/hydrodynamic shapes like aircraft wings, turbine blades, car bodies, and hulls. The learned model maps a compact set of design variables to 3D surface points representing the shape. We ensure the surface smoothness and continuity of generated geometries by incorporating an FFD layer into the generative model. We demonstrate FFD-GAN's performance using a wing shape design example. The results show that FFD-GAN can generate realistic designs and form a reasonable parameterization. We further demonstrate FFD-GAN's high representation compactness and capacity by testing its design space coverage, the feasibility ratio of the design space, and its performance in design optimization. We demonstrate that over 94\% feasibility ratio is achieved among wings randomly generated by the FFD-GAN, while FFD and B-spline only achieve less than 31\%. We also show that the FFD-GAN leads to an order of magnitude faster convergence in a wing shape optimization problem, compared to the FFD and the B-spline parameterizations.
\end{abstract}

\section{Introduction}

Airfoil design is essential for the performance of aerodynamic objects such as aircraft wings and turbine blades. An airfoil design problem search for a set of design variables that maximizes the performance. The number (dimensionality) of design variables plays an important role in the convergence and computational cost of the optimization. Particularly, gradient-free optimization (\eg, Bayesian optimization and population-based optimization) can be prohibitively expensive for high-dimensional design variables due to the curse of dimensionality~\cite{Bellman34}, especially when using a high-fidelity computational fluid dynamics (CFD) simulation. Thus, dimensionality reduction (DR) methods have been proposed to derive more compact airfoil shape representations~\cite{viswanath2011dimension,viswanath2014constrained,poole2015metric,cinquegrana2017efficient,cinquegrana2018investigation,yasong2018global,li2018data,poole2019efficient,chen2019aerodynamic,chen2020airfoil,chen2020mo}. Most research focuses on reducing two-dimensional (2D) airfoil design variables. For three-dimensional (3D) airfoils, dimensionality reduction is usually applied only to 2D airfoils and 3D geometries are formed by stacking multiple 2D airfoils~\cite{poole2019efficient}. In this paper, we address the problem of deriving a compact representation for the 3D airfoil as a whole. Specifically, we propose a deep generative model, FFD-GAN, which learns a lower-dimensional \textit{latent space} from a 3D airfoil database. It has the following advantages:
\begin{enumerate}
\item Comparing to commonly used linear dimensionality reduction such as singular value decomposition (SVD)~\cite{poole2019efficient} and principal component analysis (PCA)~\cite{berguin2015dimensionality}, the FFD-GAN can model much more complex nonlinear correlation in the data, and hence can reach higher representation compactness. Also, its \textit{FFD layer} uses the Free-Form Deformation (FFD) parameterization to ensures the smoothness and continuity of generated airfoil surfaces;
\item FFD-GAN is trained to only generate realistic designs as it learns the distribution of real-world examples, whereas traditional parameterizations (\eg, free-form deformation (FFD)~\cite{kenway2010cad}, and B-spline surfaces~\cite{hicken2010aerodynamic}, and class and shape transformation (CST)~\cite{kulfan2008universal}) or non-generative model-based dimensionality reduction techniques do not prohibit generating invalid designs.
\end{enumerate}

The FFD-GAN contains a generator that maps latent variables to 3D shapes (represented by surface points), we can treat the latent variables and the mapping as design parameters and the parametric function, respectively. Thus, in a broader sense, FFD-GAN provides a new parameterization which is inferred from data, rather than fixed apriori as in common sense.
As a result, we can use FFD-GAN for 1)~more efficient design space exploration and design optimization; and 2)~generating new designs to augment the 3D airfoil database for the purpose of improving the accuracy of CFD surrogate models.

In this paper, we describe the details of the proposed FFD-GAN, and use a wing design example to demonstrate two properties of the representation learned by FFD-GAN: 1)~\textit{representation capacity}, \ie, the coverage of the design space; and 2)~\textit{representation compactness}, \ie, the ability to use the least design variables to cover a sufficient design space and excluding invalid (unrealistic) designs.
The representation capacity determines the theoretical bounds of the optimal performance that an aerodynamic shape optimization can achieve; while higher representation compactness leads to lower computational cost and faster convergence in practice~\cite{poole2019efficient,chen2019aerodynamic,chen2020airfoil,chen2020mo}. A highly compact representation is also favorable in design space exploration and can be used for generating new designs. We demonstrate that FFD-GAN supports both properties by testing the representation's design space coverage, feasibility ratio, and its efficiency in shape optimization.

Our main contributions are listed as follows:
\begin{enumerate}
\item We propose a new deep generative model-based 3D aerodynamic shape parameterization method, FFD-GAN, which learns the parameterization from data;
\item We design a series of experiments (\eg, latent traversal, fitting test, feasibility ratio test, and shape optimization) that can be used for comparing the representation capacity and compactness of parameterizations;
\item We demonstrate that FFD-GAN significantly improves the representation compactness upon commonly used parameterizations like FFD and B-spline surface, while having comparable representation capacity;
\item We describe a probabilistic grammar for automatically creating a wing shape dataset containing 47,344 designs. This can facilitate the future study of data-driven aircraft wing design.
\end{enumerate}

\section{Background}

In this section, we introduce previous work on reducing the design space dimensionality of aerodynamic designs (Sec.~\ref{sec:dr}) and deep learning-based 3D shape synthesis (Sec.~\ref{sec:3d_shape_synth}). We also briefly describe the deep generative model that our proposed model is built on (Sec.~\ref{sec:wgan_gp}).

\subsection{Design Space Dimensionality Reduction}
\label{sec:dr}

It is usually wasteful to search for optima in the design spaces of normal parameterizations (\eg, FFD and B-spline surface), since valid designs only constitute a small portion of those spaces so that most CFD evaluations are performed on invalid designs. Past work has studied methods to obtain more compact representations via dimensionality reduction. Factor screening methods~\cite{welch1992screening,myers1995response} can select the most relevant design variables for a design problem while fixing the rest as constant during optimization. These methods fail to consider the correlation between design variables. In response, researchers have studied ways to capture the low-dimensional subspace that identifies important directions with respect to the change of \textit{response} (\ie, QoI or performance measure)~\cite{lukaczyk2014active,berguin2014dimensional,berguin2014dimensionality,berguin2015method,grey2018active}. This response-based dimensionality reduction usually has several issues: 
(1)~it requires many simulations when collecting samples of response gradients;
(2)~variation in gradients can only capture nonlinearity rather than variability in the response, so extra heuristics are required to select latent dimensions that capture steep linear response changes;
(3)~the learned latent space is not reusable for any different design space exploration or optimization task (\ie, when a different response is used); and 
(4)~the linear DR techniques applied in previous work may not model well responses with a nonlinear correlation between partial derivatives.

The first three issues can be avoided by directly applying DR on design variables without associating them with the response. Doing so assumes that if changes in a design are negligible, changes in the responses are also negligible. In the area of aerodynamic design, researchers use linear models such as principal component analysis (PCA)~\cite{cinquegrana2017efficient,cinquegrana2018investigation,yasong2018global} or singular value decomposition (SVD)~\cite{li2018data,allen2018wing,poole2019efficient} to reduce the dimensionality of design variables. 
Although those linear models provide optimal solutions to the linear DR problem, their linear nature makes them unable to achieve the most compact representation (\ie, use the least dimensions to retain similar variance in the data) when the data is nonlinear, which is the most common case for real-world data.
Nonlinear models like generative topographic mapping~\cite{viswanath2011dimension,viswanath2014constrained} can solve this problem to some extent, but are still limited to the assumption that data follow a Gaussian mixture distribution, which is too strict in most real-world cases. Beyond these data-driven methods, genetic modal design variables (GMDV)~\cite{kedward2020towards} generates airfoils through orthogonal modes derived from the reduced singular matrix of the third-difference matrix. None of these DR methods encourage compactness of the reduced shape representation, where the volume of the latent space that maps to the domain of invalid designs are minimized.
Complementary work in DR has been done in other fields such as computer vision and computer graphics~\cite{lee2007nonlinear,goodfellow2016deep}, where DR is used for generating images or 3D shapes. Deep generative networks such as variational autoencoders (VAEs)~\cite{kingma2013auto} and generative adversarial networks (GANs)~\cite{goodfellow2014generative} have been widely applied in those areas to learn the latent data representations. These methods are known for their ability to learn a compact latent representation from complex high-dimensional data distributions, where the latent representation follows a simple, known distribution (\eg, a normal or uniform distribution).
Our work extends this class of techniques by considering the generation of smooth geometries such as those needed in spline-based representations.

\subsection{Learning 3D Shape Synthesis}
\label{sec:3d_shape_synth}

In this work, we want to learn a model that generates/parameterizes realistic 3D aerodynamic shapes. Previous researches have taken various approaches that synthesize 3D objects by learning from a shape database. Neural network-based models such as multilayer perceptrons (MLPs), VAEs, and GANs are commonly used~\cite{wu2016learning,wang2018global,richter2018matryoshka,tatarchenko2017octree,achlioptas2018learning,arsalan2017synthesizing,ben2018multi,sinha2017surfnet,chen2019learning,park2019deepsdf}. While those models can generate realistic 3D shapes with high visual quality, it is problematic to directly use them to create aerodynamic shapes. Because there is no guarantee that the shape produced by the model has a smooth and continuous surface, which is an important requirement for aerodynamic performance. We solve this problem by using an FFD layer to ensure the surface smoothness and continuity of generated shapes. The concept of the FFD layer has been used in 3D shape reconstruction tasks~\cite{jack2018learning,kurenkov2018deformnet}, where the objective is to predict a shape that best matches the shape in a given image. The FFD layer is used for deforming a template shape to match the input image. In this work, we address the problem of 3D shape generation, where we learn the distribution of 3D airfoil shapes, so that we can sample realistic shapes via a compact set of parameters (\ie, latent variables). Here we use the FFD layer for the consideration of the shape's aerodynamic performance.

\subsection{Wasserstein Generative Adversarial Network with Gradient Penalty}
\label{sec:wgan_gp}

The proposed FFD-GAN learns the distribution of 3D airfoil shapes by using a generative adversarial network (GAN)~\cite{goodfellow2014generative}. A vanilla GAN has two components: 1)~a generator $G$ that generates a sample $\Tilde{\bm{X}}\in \mathbb{R}^{D}$ given any \textit{noise vector} $\bm{z}\in \mathbb{R}^{d_{\bm{z}}}$ drawn from a prior distribution $P_{\bm{z}}$, \ie, $\Tilde{\bm{X}}=G(\bm{z})$; and 2)~a discriminator $D$ acting as a classifier that distinguishes whether any given samples are from the real database (\ie, real samples) or generated by the generator (\ie, fake samples). Both components improve during training via a minimax optimization, \ie, $D$ minimizes the classification error and $G$ maximizes the chance of a generated sample being misclassified. This objective can be expressed as
\begin{equation}
\min_G\max_D V(D,G) = \mathbb{E}_{\bm{X}\sim P_{\mathrm{data}}}[\log D(\bm{X})] + \mathbb{E}_{\bm{z}\sim P_{\bm{z}}}[\log(1-D(G(\bm{z})))],
\label{eq:gan}
\end{equation}
where $P_\mathrm{data}$ denotes the data distribution.

Once trained properly, $G$ can convert any random latent vector $\bm{z}\sim P_{\bm{z}}$ to a new sample that resembles those in the database. Thus, essentially $G$ can be considered as a parameterization for the generated shape $\Tilde{\bm{X}}$ (by using the latent vector $\bm{z}$ as shape parameters or design variables). Also, since the neural networks can learn a complex and highly nonlinear mapping from $\bm{z}$ to $\Tilde{\bm{X}}$, we can make the space of $\bm{z}$ (\textit{latent space}) as compact as possible and hence serve the purpose of nonlinearly reducing the dimensionality of the original design variables (\ie, by making $d_{\bm{z}}\ll D$). 

During experiments, we found that the vanilla GAN's training process is unstable. This may be caused by the fact that the divergence between the data distribution and the generator distribution (\ie, the distribution of generated samples) that vanilla GAN is minimizing are not continuous with respect to the generator's parameters~\cite{arjovsky2017wasserstein}. Therefore, we use a Wassersterin GAN with gradient penalty (WGAN-GP)~\cite{gulrajani2017improved} to avoid this problem. It uses the Earth-Mover (Wasserstein-1) distance between the data distribution and the distribution of generated samples. Under mild assumptions, the Earth-Mover distance is continuous and differentiable almost everywhere. The objective becomes
\begin{equation}
\min_G\max_{D\in \mathcal{D}} W(D,G) = \mathbb{E}_{\bm{X}\sim P_{\mathrm{data}}}[D(\bm{X})] - \mathbb{E}_{\bm{z}\sim P_{\bm{z}}}[D(G(\bm{z}))],
\label{eq:wgan}
\end{equation}
where $\mathcal{D}$ is the set of 1-Lipschitz functions. To enforce the Lipschitz constraint on the discriminator, WGAN-GP constraints the gradient norm of the discriminator’s output with respect to its input. This is achieved by adding the following gradient penalty term to the objective:
\begin{equation}
R_1(D) = \mathbb{E}_{\hat{\bm{X}}\sim P_{\hat{\bm{X}}}}\left[ \left( \left\| \triangledown_{\hat{\bm{X}}}D(\hat{\bm{X}}) \right\|_2-1 \right)^2 \right],
\label{eq:gp}
\end{equation}
where $\hat{\bm{X}}$ is sampled uniformly along straight lines between pairs of points sampled from the data distribution and the generator distribution. The final objective of WGAN-GP then becomes
\begin{equation}
\min_G\max_D W(D,G) - \lambda_1 R_1(D),
\label{eq:wgan_gp}
\end{equation}
where $\lambda_1$ is the weight of the gradient penalty term. 


Although WGAN-GP mitigates GAN's training stability issue, it does not have any mechanism to constrain the surface smoothness of generated shapes. Our proposed FFD-GAN adapted the WGAN-GP model and further solves the issue of generating smooth surfaces by using an FFD layer in the generator.

\section{FFD-GAN: Deep Generative Model-based 3D Airfoil Parameterization}

In this section, we describe the detailed model architecture and the objective of our proposed FFD-GAN model.

\subsection{Model Architecture}

\begin{figure}[hbt!]
\centering
\includegraphics[width=1.0\textwidth]{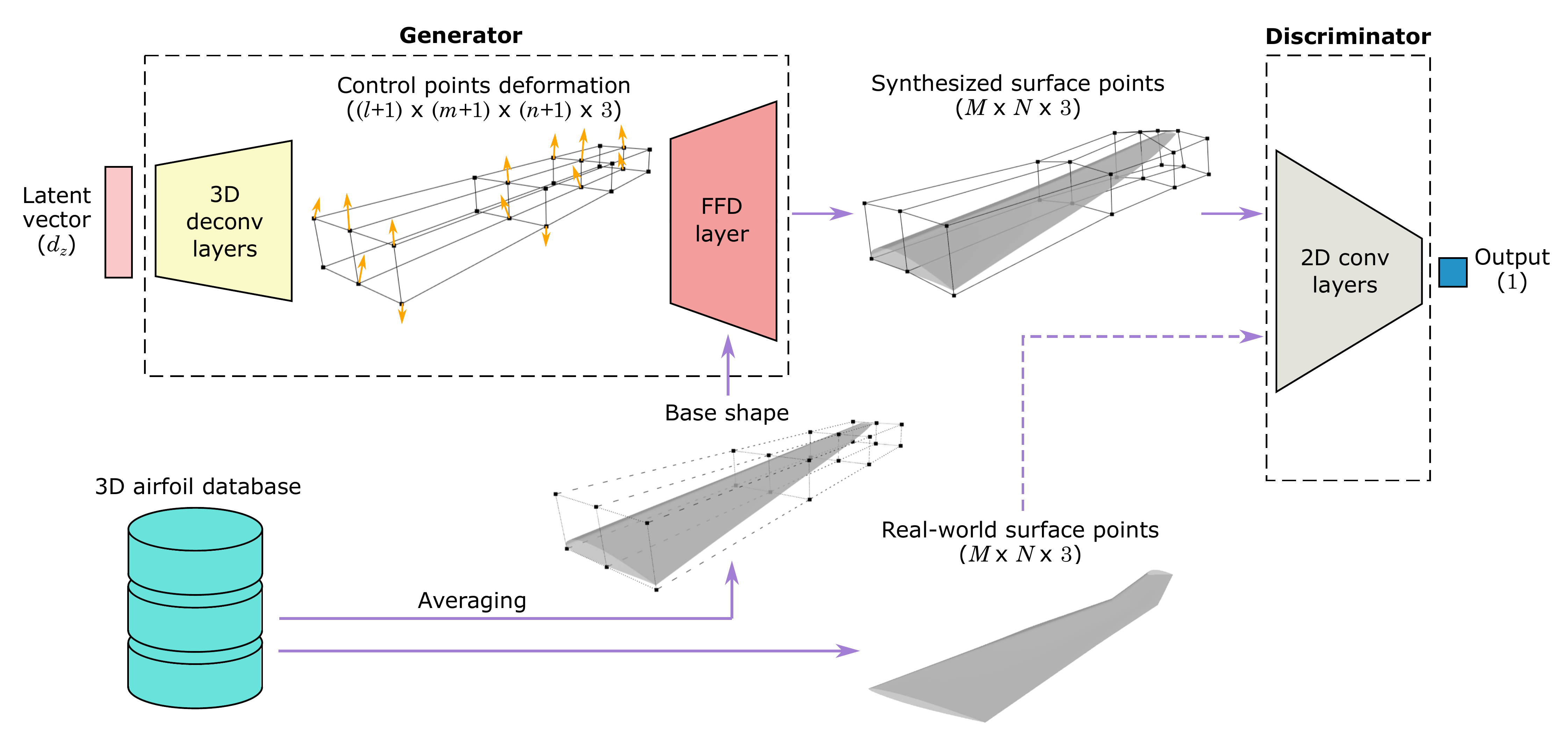}
\caption{Architecture of FFD-GAN.}
\label{fig:architecture}
\end{figure}

The architecture of FFD-GAN is shown in Fig~\ref{fig:architecture}. The FFD-GAN uses its generator to produce parameters that deform a base shape to new designs. The base shape can be obtained by averaging all shapes from a design database. We can compute a set of equally distributed FFD control points $\bm{P}^{\mathrm{base}}$ correspond to this base shape~\cite{sederberg1986free,masters2017geometric}:
\begin{equation}
\bm{P}^{\mathrm{base}}_{i,j,k} = \left( x^{\mathrm{base}}_{\mathrm{min}}+\frac{i}{l}(x^{\mathrm{base}}_{\mathrm{max}}-x^{\mathrm{base}}_{\mathrm{min}}), y^{\mathrm{base}}_{\mathrm{min}}+\frac{j}{m}(y^{\mathrm{base}}_{\mathrm{max}}-y^{\mathrm{base}}_{\mathrm{min}}), z^{\mathrm{base}}_{\mathrm{min}}+\frac{k}{n}(z^{\mathrm{base}}_{\mathrm{max}}-z^{\mathrm{base}}_{\mathrm{min}}) \right),
\end{equation}
where $i=0,...,l$, $j=0,...,m$, and $k=0,...,n$. Thus, we have $l+1$, $m+1$, and $n+1$ control points along the $x$, $y$, and $z$ axis, respectively.

The generator consists of several 3D deconvolutional layers and an FFD layer. These 3D deconvolutional layers~\cite{choy20163d} convert a $d_z$-dimensional latent vector $\bm{z}$ into $\Delta\bm{P}$, an $(l+1)\times (m+1)\times (n+1)\times 3$ tensor representing the deformation of $(l+1)\times (m+1)\times (n+1)$ FFD control points. Both $\Delta\bm{P}$ and $\bm{P}^{\mathrm{base}}$ are then fed into the FFD layer to produce the set of $M\times N$ 3D airfoil surface points. Specifically, $\Delta\bm{P}$ and $\bm{P}^{\mathrm{base}}$ are first added to form the control points $\bm{P}$ for the deformed shape. The surface points $\Tilde{\bm{X}}$ are then computed by the following~\cite{sederberg1986free,masters2017geometric}:
\begin{equation}
\Tilde{\bm{X}}(u,v,w)=\sum_{i=0}^{l}\sum_{j=0}^{m}\sum_{k=0}^{n}B_i^l(u)B_j^m(v)B_k^n(w)\bm{P}_{i,j,k},
\label{eq:ffd}
\end{equation} 
where $0\leq u\leq 1$, $0\leq v\leq 1$, and $0\leq w\leq 1$ are parametric coordinates, and the $l$-degree Bernstein polynomials $B_i^l(u)=\binom{l}{i}u^i(1-u)^{l-i}$. We set the parametric coordinates based on the surface points of the base shape:
\begin{equation} 
(\bm{u}, \bm{v}, \bm{w}) =
\left(
\frac{\bm{x}^{\mathrm{base}}-x^{\mathrm{base}}_{\mathrm{min}}}{x^{\mathrm{base}}_{\mathrm{max}}-x^{\mathrm{base}}_{\mathrm{min}}}, 
\frac{\bm{y}^{\mathrm{base}}-y^{\mathrm{base}}_{\mathrm{min}}}{y^{\mathrm{base}}_{\mathrm{max}}-y^{\mathrm{base}}_{\mathrm{min}}},
\frac{\bm{z}^{\mathrm{base}}-z^{\mathrm{base}}_{\mathrm{min}}}{z^{\mathrm{base}}_{\mathrm{max}}-z^{\mathrm{base}}_{\mathrm{min}}}
\right).
\end{equation}

Equation~(\ref{eq:ffd}) is differentiable with respect to $\Delta\bm{P}$ and hence it allows gradients to be backpropagated to the previous layers.

\begin{figure}[hbt!]
\centering
\includegraphics[width=1.0\textwidth]{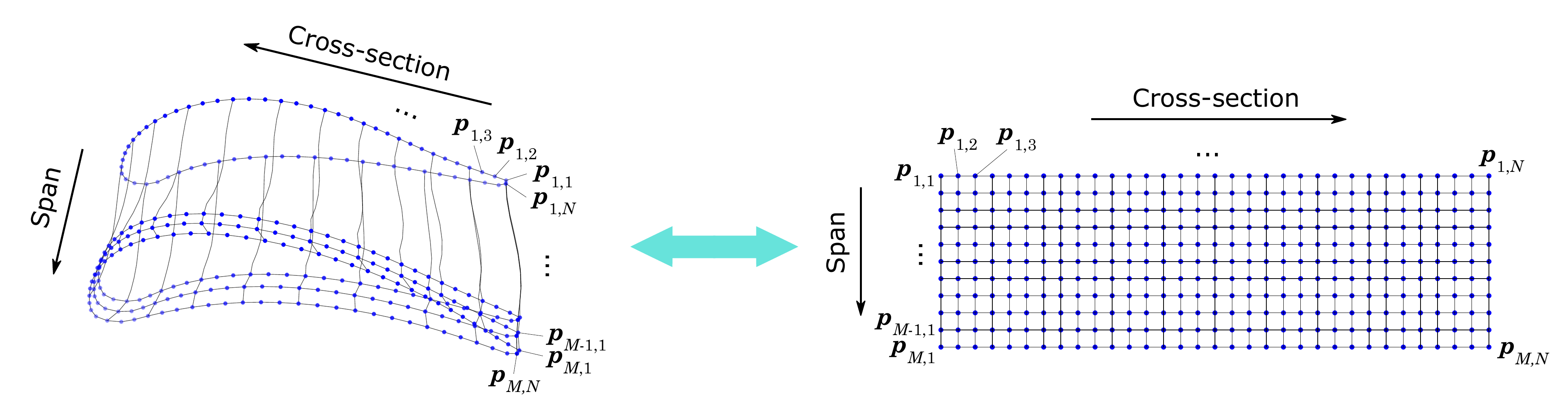}
\caption{3D airfoil surface points are represented as an $M\times N\times3$ tensor. Here $\bm{p}_{s,t}\in \mathbb{R}^3$ denotes the 3D coordinate at the $(s,t)$-th surface point.}
\label{fig:points}
\end{figure}

The discriminator $D$ takes the surface points as input and outputs a scalar which is then used in Eq.~(\ref{eq:wgan_gp}) to compute the Earth-Mover distance and the gradient penalty. Specifically, as shown in Fig.~\ref{fig:points}, the surface points are on a grid of $M\times N$, where $M$ is the number of cross-sections along the span and $N$ is the number of surface points on each section. The coordinates of these surface points are thus represented as a tensor with shape $M\times N\times3$ (equivalent to a three-channel image with a resolution of $M\times N$).

The trained generator can form a parametric model which maps any $d_z$-dimensional latent vector $\bm{z}$ to a set of surface points representing a smooth 3D shape. Note that after training, the discriminator is discarded and only the generator is used for parameterizing the shapes. The WGAN-GP objective in Eq.~(\ref{eq:wgan_gp}) ensures that generated 3D shapes are drawn from a distribution close to the data distribution. Therefore, the (low-dimensional) latent vector can represent shape variations in the dataset. This provides us with a new compact parameterization with sufficient representation capacity to cover existing designs in the training dataset.

\subsection{Regularization}

\begin{figure}[hbt!]
\centering
\includegraphics[width=1.0\textwidth]{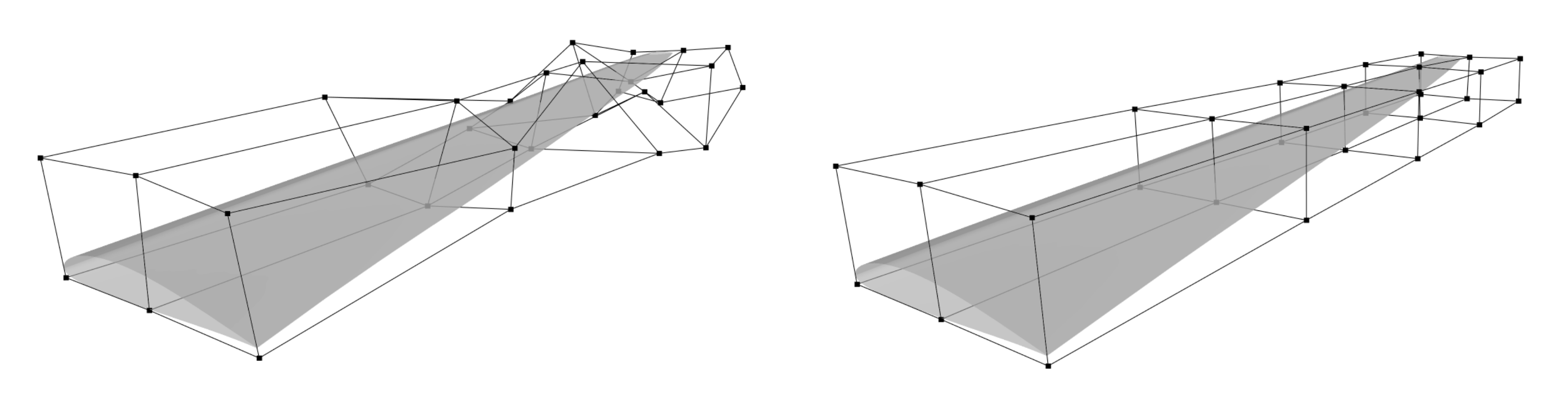}
\caption{The same shape with different FFD parameterizations.}
\label{fig:regularization}
\end{figure}

The FFD representation (\ie, the choice of $\Delta\bm{P}$) for a set of surface points is not unique. As shown in Fig.~\ref{fig:regularization}, the same shape can have different FFD parameterizations. To limit possible solutions of $\Delta\bm{P}$ and make the generator converge to an optimum with a reasonable FFD representation, we add the following regularization: 
\begin{equation}
R_2(G) = \frac{1}{B(l+1)(m+1)(n+1)}\sum_{s=1}^B\sum_{i=0}^l\sum_{j=0}^m\sum_{k=0}^n \big\|\Delta \bm{P}_{i,j,k}^{(s)}\big\|^2,
\label{eq:reg_deltap}
\end{equation}
where $B$ is the batch size.

With the above regularization term, the loss function of FFD-GAN becomes 
\begin{equation}
\min_{G}\max_D W(D,G) - \gamma_1 R_1(D) + \gamma_2 R_2(G),
\label{eq:loss}
\end{equation}
where $\gamma_2$ is the weight of the regularization term. The first two terms, $W(D,G)$ and $R_1(D)$, are from Eq.~(\ref{eq:wgan_gp}).

\section{Test Results on a Wing Design Example}

In this section, we use a wing design example to demonstrate the performance of the FFD-GAN as a new parameterization, and compare it with two common parameterizations~\textemdash~FFD and B-spline surface.

\subsection{Dataset Creation and Preprocess}

\begin{figure}[hbt!]
\centering
\includegraphics[width=1.0\textwidth]{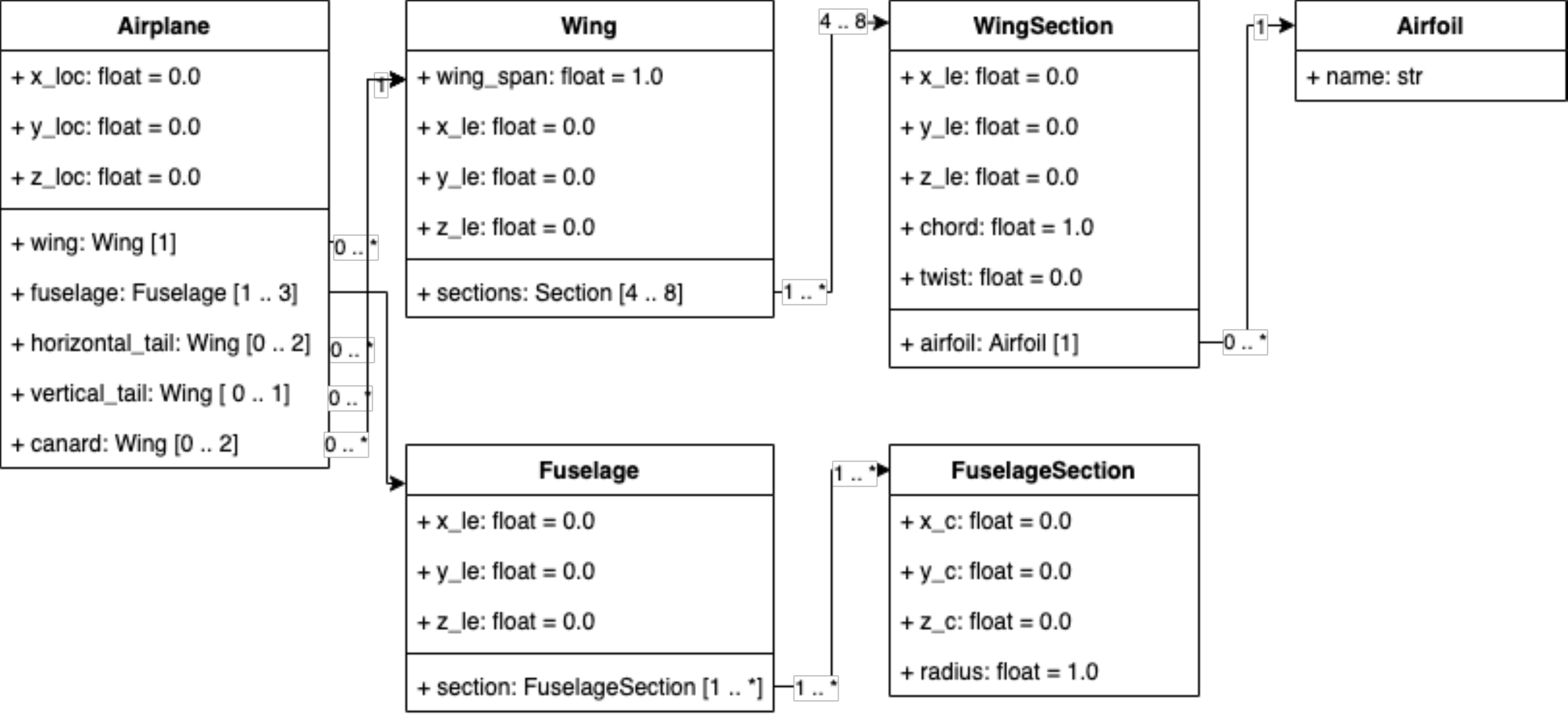}
\caption{An illustration of the grammar for the airplane from which the probabilistic grammar for the wing is generated.}
\label{fig:grammar_airplane}
\end{figure}

To create a dataset for the FFD-GAN to learn from, we sample wings from a comprehensive design space in which realistic wing shape variations are realized. We represent the entire design space in the form of a probabilistic grammar~\cite{kar2019metasim} that represents the ontology of the generated design. An example of one grammar (although, not probabilistic) representing an aircraft is shown in Fig.~\ref{fig:grammar_airplane}, where each aircraft is comprised of $1\sim 3$ fuselages, a wing, $0\sim 2$ horizontal tails, 0 or 1 vertical tail, and $0\sim 2$ canards. Further, the fuselage is parameterized by a set of one or more sections, with each section being represented by 4 parameters~\textemdash~the circumferential radius (assuming tubular fuselage) and the coordinate offsets. Similarly, the wing (which is symmetric) is parameterized by its span and a set of $4\sim 8$ sections defining the profile of the wing. A minimum of four sections are chosen to ensure sufficient definition in the wing, while a maximum of eight are chosen as it was observed that having greater than eight sections would result in too frequent changes to the wing and lead to unrealistic shapes. Each section is further parameterized by its location, chord length, and twist along with the airfoil defining the section.

\begin{figure}[hbt!]
\centering
\includegraphics[width=1.0\textwidth]{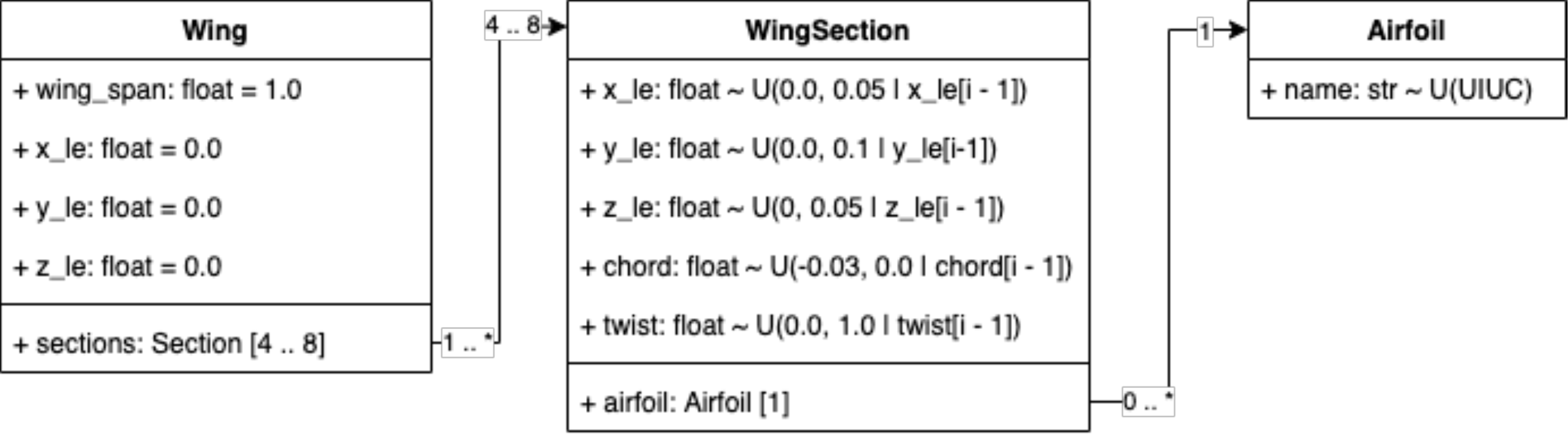}
\caption{Probabilistic grammar associated with the data synthesis process.}
\label{fig:prob_grammar}
\end{figure}

Based on this parameterization of the aircraft, a probabilistic grammar is constructed for the wing as shown in the Fig.~\ref{fig:prob_grammar}. The parameters of any wing section are sampled based on the parameters of the previous section. Mathematically, if $P_{\mathrm{wing}}$ is the probabilistic grammar representing the design space of the wings, we can represent $w$ as an instance of the wing sampled from the design space, \ie, $w \sim P_{\mathrm{wing}}$. Due to the hierarchical nature of the representation, we can utilize a graph to represent each instance. This graph is given as $G$ such that $G = (V, E, A)$, where $V$, $E$, and $A$ are the sets of nodes, edges (\ie, compositional relationship), and attributes associated with the nodes, respectively. Thus, a procedural generation process is implemented in which we realize the following sequence:
\begin{itemize}
    \item num. of sections $\sim \mathcal{U}\{4, 8\}$,
    \item airfoil[section $i$] $\sim \mathcal{U}\{\text{UIUC Database}\}$,
    \item attribute[section $i$] $\sim \mathcal{U}(\text{LB}, \text{UB} | \text{attribute}[\text{section } (i-1)])$,
\end{itemize}
where LB and UB denotes the lower and the upper bounds of the attribute, respectively.

\begin{figure}[hbt!]
\centering
\includegraphics[width=1\textwidth]{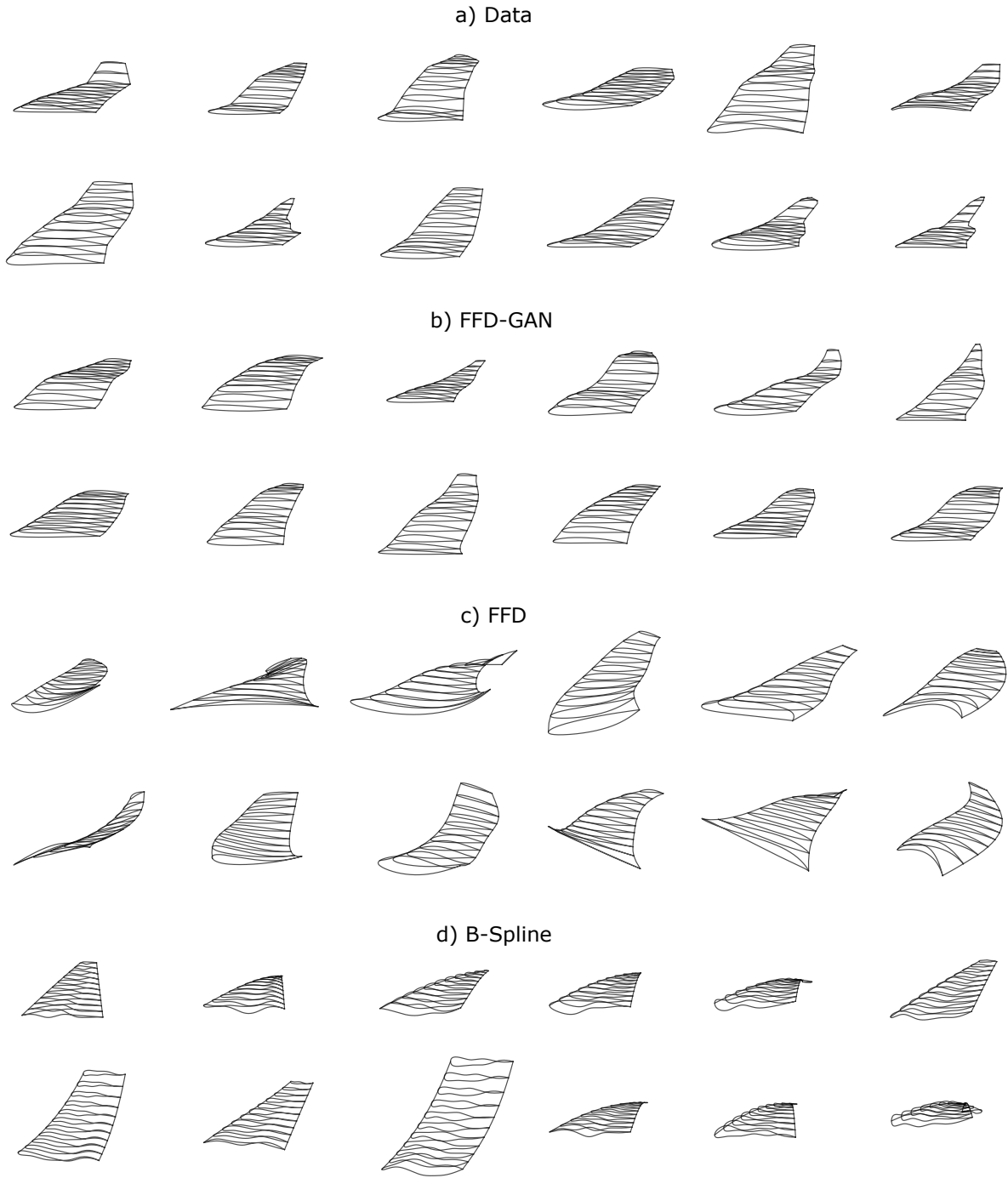}
\caption{Comparison of wings randomly sampled or generated from a)~the database, b)~FFD-GAN with $d_{\bm{z}}=15$, c)~FFD with $3\times 4\times 2$ control points, and d)~B-spline surface with $4\times 14$ control points. The 2D plots show wing cross-sections in the xz view.}
\label{fig:samples}
\end{figure}

This implies that the number of sections in each wing are uniformly sampled between 4 and 8. We linearly interpolate the coordinates of the wing between sections to form a smooth transition. At each section, the chord length, the sectional twist, and the leading edge position ($x$ and $z$ dimensions) are sampled based on a known distribution. Each section profile is randomly sampled from the UIUC database containing 1,550 predefined airfoils (ranging from 2032c to ys930)\footnote{\url{https://m-selig.ae.illinois.edu/ads/coord_database.html}}. Thus, the entire wing is parameterized by a set of $20 \sim 40$ parameters (varying depending on the number of sections chosen). 
The probabilistic grammar introduces a set of rules to ensure a valid wing section is sampled. For example, to ensure a smooth transition at the leading edge, an incremental design space is chosen: the parameters of section $i + 1$ is dependent on those of section $i$. The probabilistic grammar parameterizes the rules themselves using uniform distributions as illustrated in Figure \ref{fig:prob_grammar}. For example, the variation of the chord from root to tip has to follow a monotonic variation, but the variation itself is not prescribed. With such rules for each parameter across all the sections, a Monte Carlo sampling is used to create new designs. This process results in a dataset that contains 47,344 feasible wing designs. We interpolate those shapes so that each of them is represented by a $21\times 199\times 3$ tensor, where 21 is the number of sections and each section has 199 surface points. We use 80\% designs as the training data and the rest as test data. Figure~\ref{fig:samples} shows some examples from the training data.

We perform the following preprocessing steps to align wing shapes in the dataset before training the FFD-GAN:
\begin{enumerate}
\item Translate the wings so that the leading edge of the first section is at the origin;
\item Rotate the wings about the axis perpendicular to the sections so that the chord line of the first section aligns with the vector $(1, 0, 0)$;
\item Scale the wings along $y$ axes so that the half span is 1.
\end{enumerate}

\subsection{Model Configurations and Experimental Settings}

We set $\gamma_1=10$ according to \cite{gulrajani2017improved}. For $\gamma_2$, we simply set it to 1. The generator takes a $d_z$-dimensional latent vector as the input and has two fully connected (FC) layers followed by four 3D deconvolutional layers. The two FC layers have 1024 and 256 units, respectively. The 3D deconvolutional layers have depths of [128, 64, 32, 1] and strides of [(2,2,2), (2,2,1), (1,2,1), (1,1,1)]. The kernel size is (4,4,4). Each FC and deconvolutional layer is followed by batch normalization and a leaky ReLU activation, except for the last deconvolutional layer. For the FFD layer, we have $(l, m, n)=(3, 7, 1)$, \ie, the number of control points at $x$, $y$, $z$ directions are 4, 8, 2, respectively. Since the $y$-coordinate of each section is fixed for the preprocessed wing shapes in the dataset, we fix the deformation in the $y$-axis and only learn the deformation in the other two axes. Note that this condition might not hold in other circumstances and FFD-GAN can learn deformation in all axes in those circumstances. The base shape is obtained by averaging shapes from the training dataset. The discriminator takes surface point coordinates as the input and has five 2D convolutional layers followed by two FC layers. The convolutional layers have depths of [32, 64, 128, 256, 512]. We use the same stride of (2,4) and the kernel size of (4,8) for all convolutional layers. The two FC layers have 1024 and 1 unit(s), respectively. Each layer in the discriminator is followed by a leaky ReLU activation, except for the last layer. During training, we set both the generator's and the discriminator's learning rate to 0.0002. We train the discriminator for five iterations per generator iteration. We run 10,000 generator iterations. The batch size is 64.

To benchmark the performance of FFD-GAN as a new parameterization for wing shapes, we compare it with two other parameterizations~\textemdash~the FFD and the B-spline surface. For FFD, we use $(l'+1)\times (m'+1)\times (n'+1)$ control points and fix the control points deformation in $y$-axis. So the number of design variables is $2(l'+1)(m'+1)(n'+1)$. Like the FFD-GAN, the base shape is also obtained by averaging shapes from the training dataset. For the B-spline surface representation, we use $(m''+1)\times (n''+1)$ control points and a degree of 3 in both directions. To reduce the number of design variables, we fix the $y$-coordinates of control points and constrain the $x$-coordinates via the leading-edge and the trailing-edge sweep angles~\cite{osusky2015drag}. We also set the section-wise starting and end control points to be the same, so that the airfoil at each section is closed at the trailing edge. The resulting parameterization has $2+(m''+1)n''$ design variables. 

We conducted a series of experiments to demonstrate the representation capacity and compactness of the three parameterizations (\ie, FFD-GAN, FFD, and B-spline surface). In particular, to allow sampling and search in the design spaces, we need to specify reasonable bounds for the design variables in those parameterizations. For FFD-GAN, we define the design space to be the $d_z$-dimensional latent space with each latent variable bounded in $[-1, 1]$. For FFD, we have $2(l'+1)(m'+1)(n'+1)$-dimensional design variables (representing the control points deformation in $x$ and $z$ axes) bounded in $[-0.1, 0.1]$. For the B-spline surface parameterization, we first fit the average shape from the training data to obtain a set of $2+(m''+1)n''$ base design variables. We then bound the leading-edge and the trailing-edge sweep angles to allow a $\pm 5\degree$ variation with respect to the base design, and bound the other design variables to allow a $\pm 0.1$ variation.

We use Athena Vortex Lattice (AVL)\footnote{\url{http://web.mit.edu/drela/Public/web/avl/}} to evaluate the aerodynamic performance of generated wings. The Mach number is fixed at 0.4 (296.4 mi/hr) and the angle of attack varies from $-5\degree$ to $10\degree$.

\begin{figure}[hbt!]
\centering
\includegraphics[width=1\textwidth]{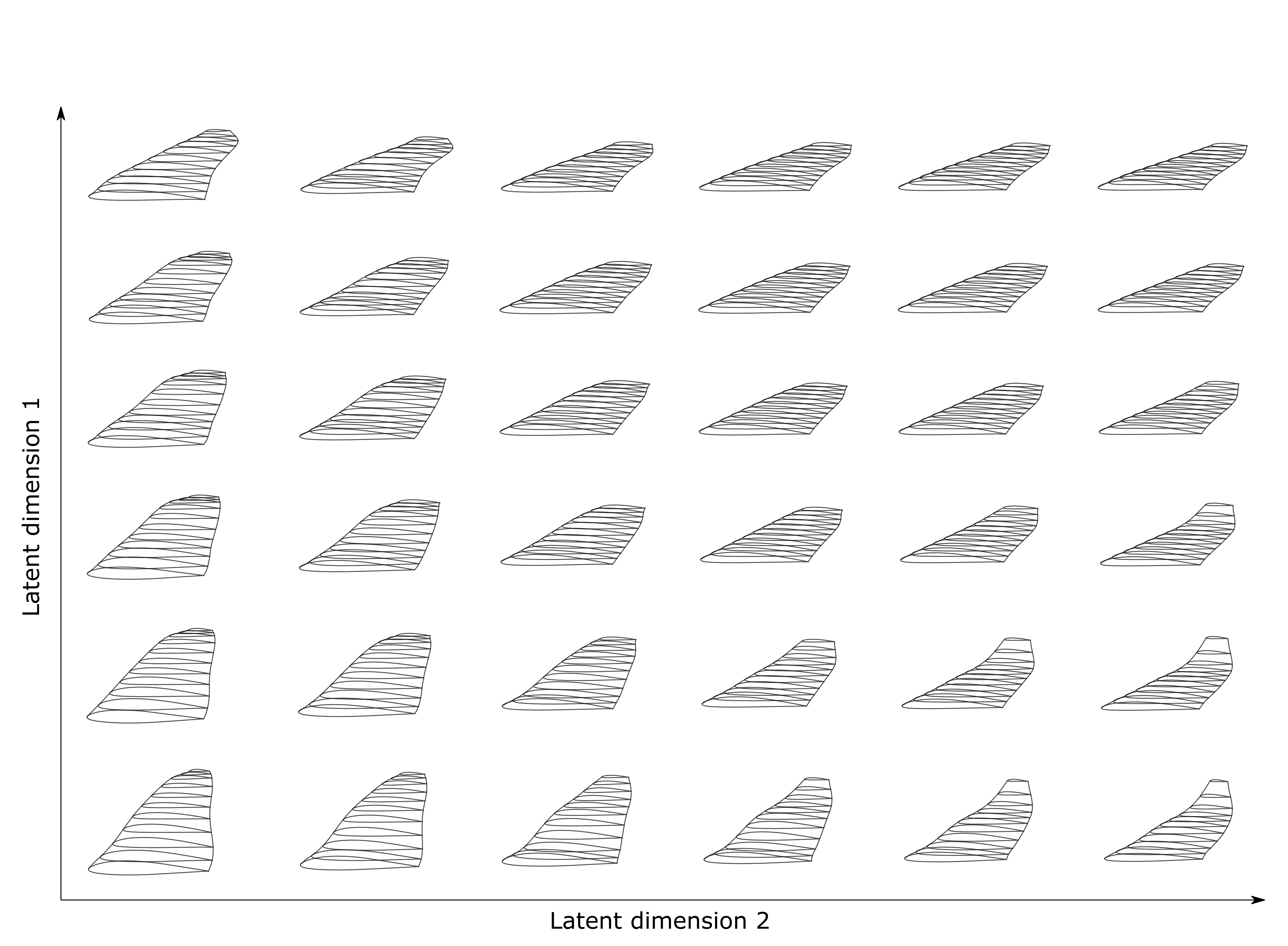}
\caption{Wings synthesized by varying the first two latent dimensions ($d_{\bm{z}}=15$).}
\label{fig:synthesized_latent_1_2}
\end{figure}

\begin{figure}[hbt!]
\centering
\includegraphics[width=1\textwidth]{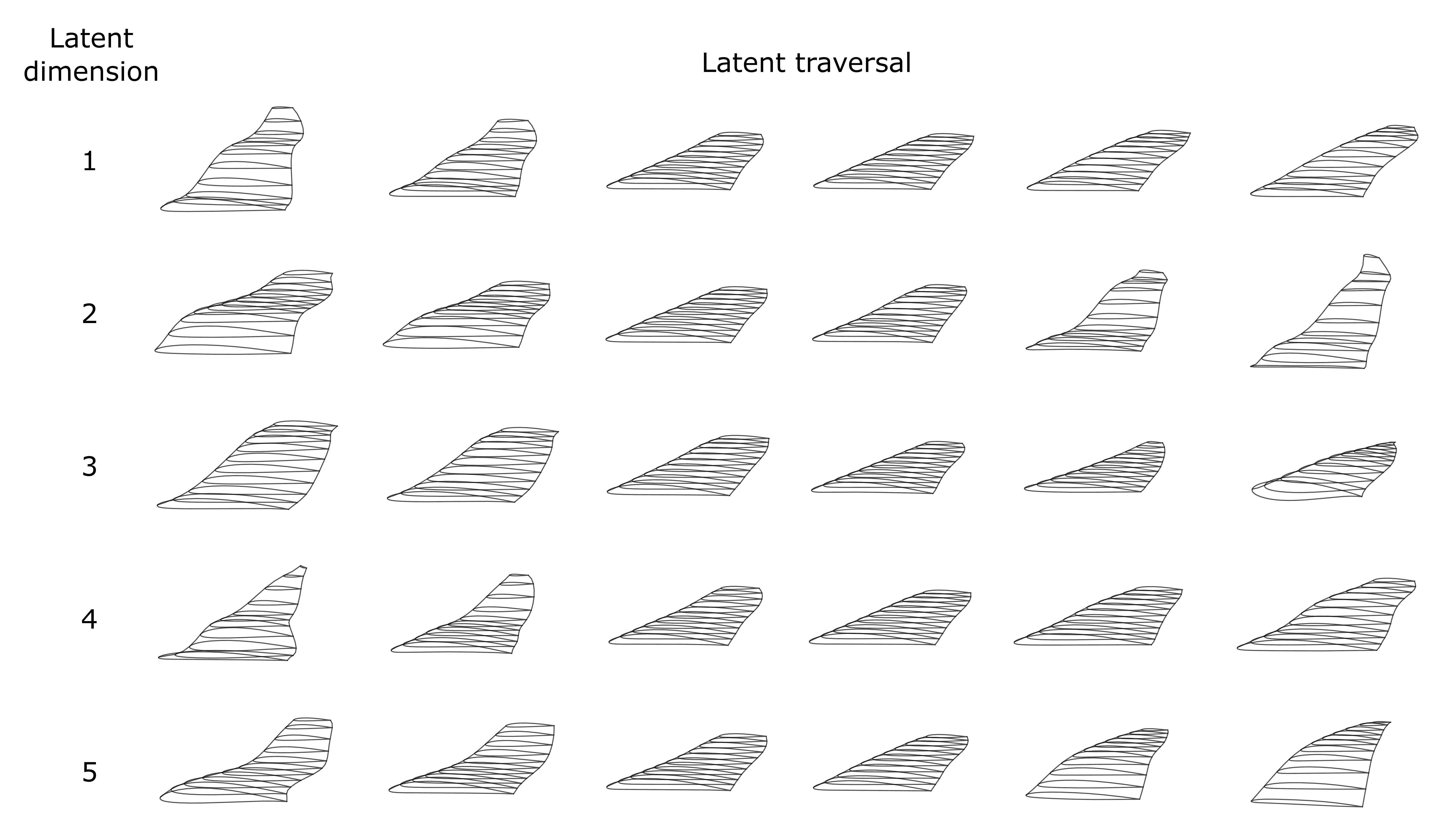}
\caption{Latent traversal of a 5-dimensional latent space. Each row is a traversal of a single latent dimension.}
\label{fig:synthesized_latent}
\end{figure}

\subsection{Wing Shape Generation}

By using a dataset of aircraft wings, we demonstrate the compactness and capacity of the representation learned by our FFD-GAN. Figure~\ref{fig:samples} shows a comparison of wings randomly sampled or generated from the dataset, FFD-GAN (FFD-GAN wings), FFD (FFD wings), and B-spline surface (B-spline wings). The FFD-GAN wings are similar to the data; while it is hard to find realistic wings in randomly generated FFD or B-spline wings. This illustrates the ability of FFD-GAN in learning the distribution of examples in the database and excluding invalid designs, which is impossible for standard parameterizations like FFD and B-spline surface. This advantage is essential in design space exploration and gradient-free shape optimization, where sampling from the design space is required. It largely reduces the computational cost for repetitively sampling and evaluating invalid designs.

Figure~\ref{fig:synthesized_latent_1_2} shows a continuous shape variation in the first two latent dimensions, which indicates a continuous transformation of shapes in the latent space. Figure~\ref{fig:synthesized_latent} shows the latent traversal of a 5-dimensional latent space, where we vary a single latent dimension while fixing the others. This further demonstrates that shapes change continuously yet differently in each dimension.

\subsection{Design Space Coverage}

We test the design space coverage of FFD-GAN, FFD, and B-spline surface. Specifically, for each parameterization, we perform least squares fitting to match the synthesized shapes with the samples randomly draw from the test dataset (target shapes). We use the following mean square error (MSE) as the objective for the least squares fitting:
\begin{equation}
\epsilon = \frac{1}{MN}\sum_{s=1}^{M}\sum_{t=1}^{N}\|\Tilde{\bm{X}}_{s,t}-\bm{X}^{\mathrm{tgt}}_{s,t}\|^2,
\label{eq:fitting_error}
\end{equation}
where $\Tilde{\bm{X}}$ is the wing's surface points synthesized by any specific parameterization and $\bm{X}^{\mathrm{tgt}}$ is the target wing's surface points. After finding the solution $\Tilde{\bm{X}}_{s,t}^*$ (fitted shape) to the least squares problem, we measure the Hausdorff distance between the fitted shape and the target one:
\begin{equation}
d_{\mathrm {H} }(\Tilde{\bm{X}}^*, \bm{X}^{\mathrm{tgt}})=\max \left\{\,\sup_{\bm{x}\in \Tilde{\bm{X}}^*}\inf_{\bm{x}'\in \bm{X}^{\mathrm{tgt}}} \|\bm{x}-\bm{x}'\|, \, \sup_{\bm{x}'\in \bm{X}^{\mathrm{tgt}}}\inf_{\bm{x}\in \Tilde{\bm{X}}^*} \|\bm{x}-\bm{x}'\|\,\right\}.
\label{eq:hausdorff}
\end{equation}

\begin{figure}[hbt!]
\centering
\includegraphics[width=1\textwidth]{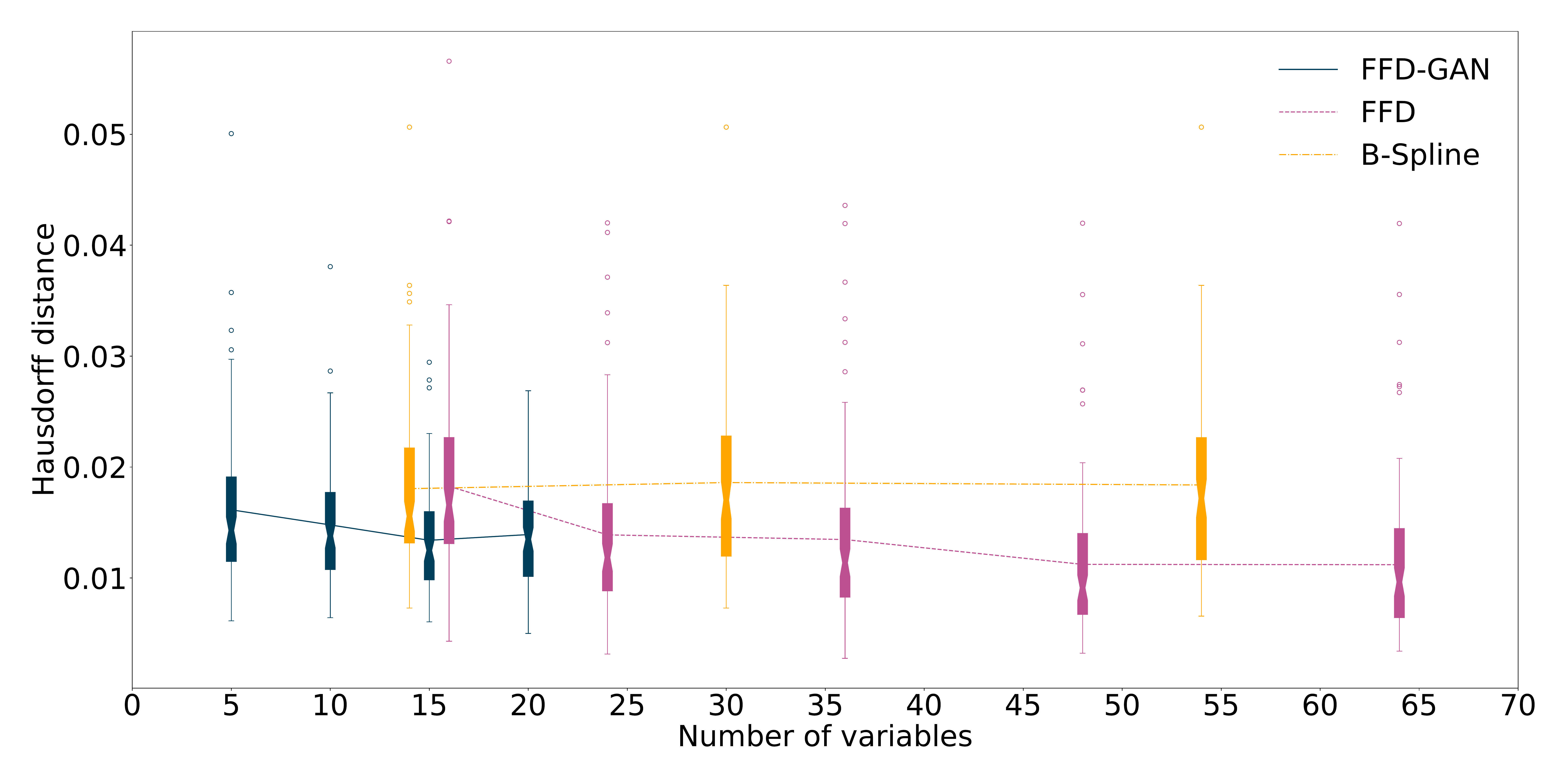}
\caption{Fitting test results showing the performance of FFD-GAN and other parameterizations in recovering the dataset. The lines connect mean values.}
\label{fig:fitting_errors}
\end{figure}

The statistics of the fitting results are shown in Fig.~\ref{fig:fitting_errors}, where we use 100 target wings randomly sampled from the database. Lower Hausdorff distances indicate a better coverage of the design space. It shows that with fewer ($<20$) design variables, FFD-GAN is better at recovering designs from data. Also, FFD-GAN with 15 design variables ($d_{\bm{z}}=15$) achieves a similar design space coverage to FFD with 24 design variables ($2\times 3\times 2$ control points). This indicates that, although FFD-GAN generates FFD parameters, its compactness is much higher than a standard FFD representation while maintaining similar representation capacity, since it needs much fewer design variables to cover a similar range of design space. This is because although the latent vector has low dimensionality, the number of control points for the FFD layer can be large enough to have a high representation capacity. The generator then learns a complex nonlinear mapping from the latent vector to the FFD parameters, allowing the compactness of the latent vector.

\begin{figure}[hbt!]
\centering
\includegraphics[width=1\textwidth]{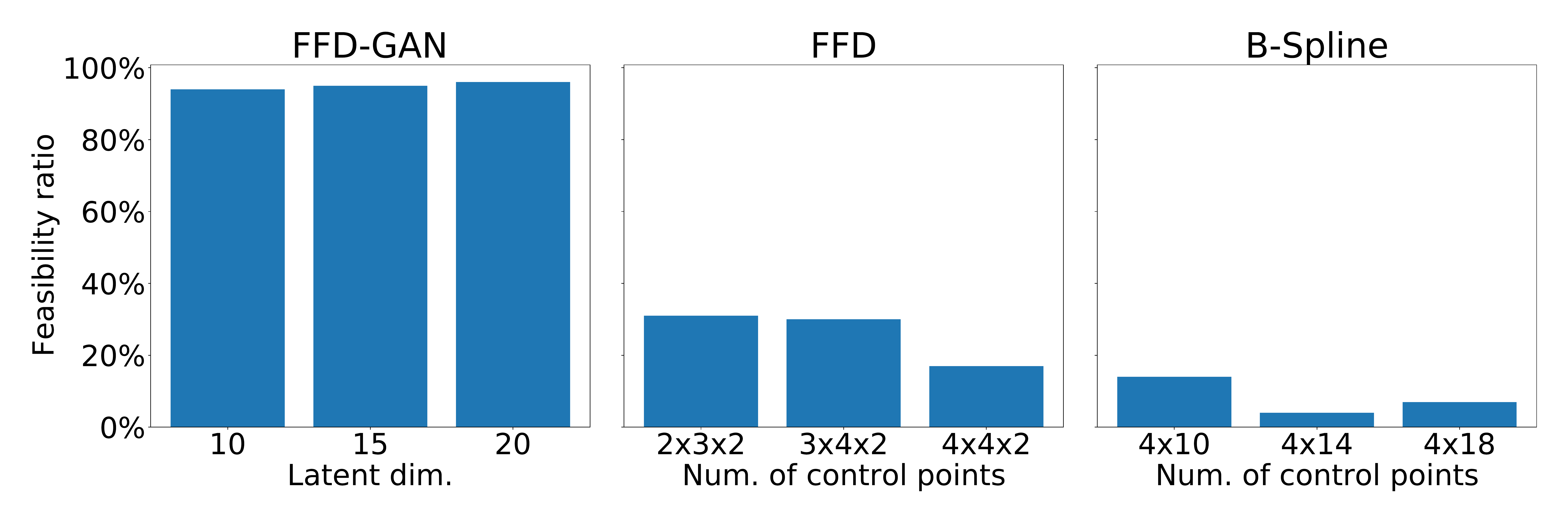}
\caption{Feasibility ratios indicated by the percentage of feasible wing designs among randomly generated samples. Here feasibility means non-self-intersecting wings with the lift to drag ratio $CL/CD>0$ at $2\degree$ angle of attack.}
\label{fig:feasibility_ratio}
\end{figure}

\subsection{Feasibility Ratio of the Design Space}

The representation compactness is also related to the volume of the design space containing valid designs (\ie, the \textit{feasibility ratio}). Thus, we use a Monte Carlo sampling to measure the percentage of feasible designs in bounded design spaces.  
The feasibility is defined by two criteria: 1)~the shape is not self-intersecting and 2)~$CL/CD>0$ at $2\degree$ angle of attack. As shown in Fig.~\ref{fig:feasibility_ratio}, FFD-GAN achieves over 94\% feasibility ratio, while FFD and the B-spline surface have less than 31\% and 14\% feasibility ratio, respectively. This indicates that the FFD-GAN forms a design space where almost everywhere in that space contains feasible designs. This is expected because the FFD-GAN's generator is trained to only generate designs that are similar to those from the existing ones, which are likely to be feasible. It automatically learns the constraints that are implicitly encoded in the data (\eg, the non-self-intersecting constraint or aerodynamic property constraints), whereas non-data-driven parameterizations cannot recognize. This causes common parameterizations like FFD and B-spline surface to represent a large number of designs violating those implicit constraints, which leads to a much lower feasibility ratio compared to the FFD-GAN.

The results on feasibility ratio and design space coverage together demonstrated the high representation compactness of FFD-GAN. Because it uses much fewer dimensions to achieve a similar or better design space coverage compared to the other two parameterizations, and is unlikely to have invalid designs in its design space.

\begin{figure}[hbt!]
\centering
\includegraphics[width=1\textwidth]{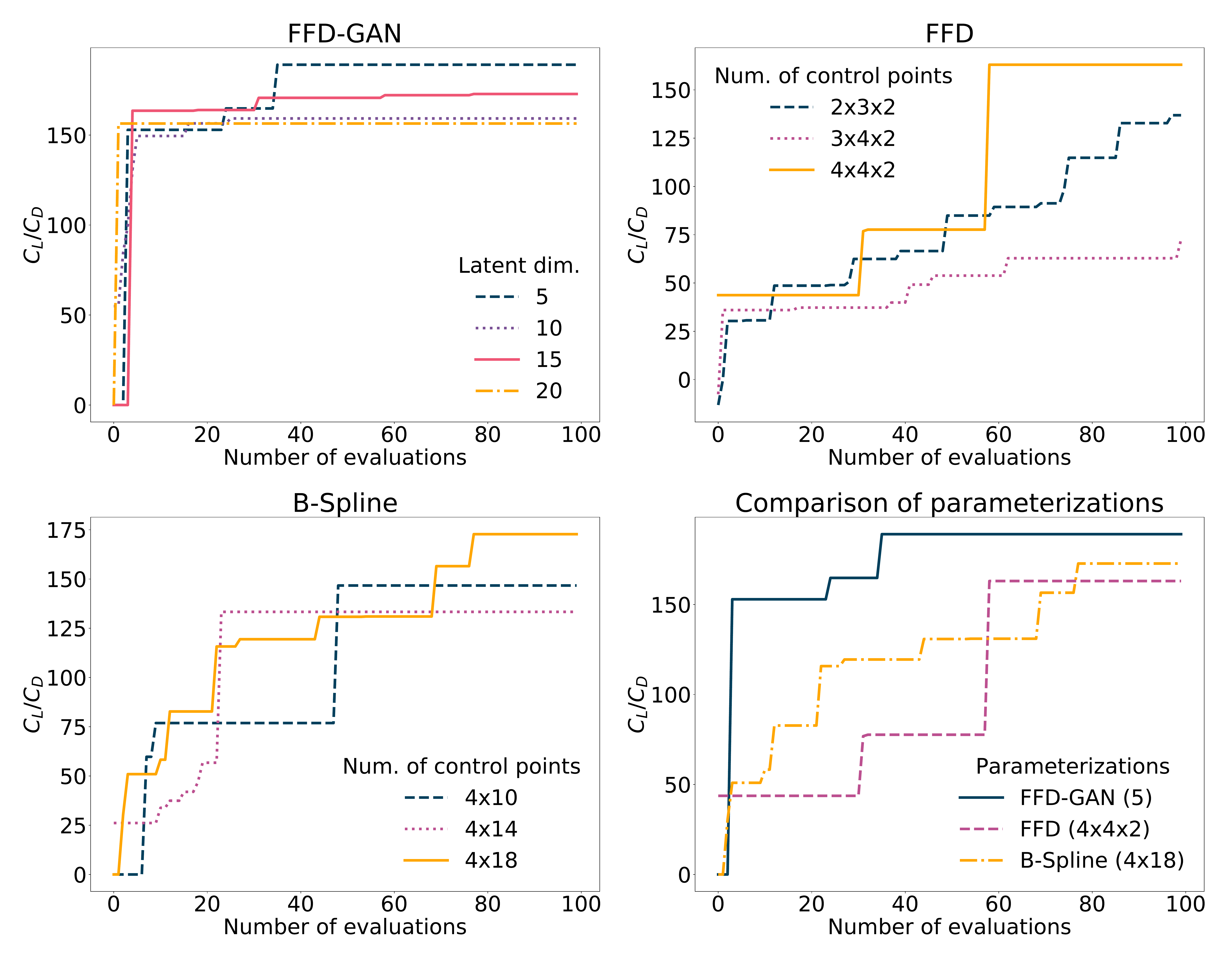}
\caption{Optimization history when using different parameterizations.}
\label{fig:opt_history_combined}
\end{figure}

\subsection{Wing Shape Optimization}

In the previous sections, we show the results for a series of tests that indicate the representation capacity and compactness improvement of the FFD-GAN over FFD and B-spline surface parameterizations. In this section, we demonstrate the benefits of such improved properties for aerodynamic shape optimization. Both representation capacity and compactness are closely related to the optimization performance: the representation capacity decides the upper bound of the optimal aerodynamic performance of represented designs, while the representation compactness affects the convergence rate. When the compactness is low, we either need to evaluate more samples to reach the global optimum due to the curse of dimensionality, or have to waste budget on evaluating a high percentage of invalid designs.

In this experiment, we maximize the lift to drag ratio $C_L/C_D$ with respect to design variables and the angle of attack (bounded in $[-5\degree, 10\degree]$). Before each evaluation, we adjust the wing's geometry such that the chord line of the first section aligns with the vector $(1, 0, 0)$ at $0\degree$ angle of attack. We use Bayesian optimization as a global optimization method that minimizes the number of evaluations. We first sample 10 initial evaluations via Latin hypercube sampling (LHS)~\cite{mckay2000comparison}. Then we use the Gaussian Process Upper Confidence Bound (GP-UCB)~\cite{srinivas2010gaussian} algorithm to sequentially suggest other 90 evaluations. Note that the optimization in this paper only provides a means to test the parameterization efficiency. The proposed FFD-GAN allows integration with any downstream optimization methods, including gradient-based ones, since we can use automatic differentiation to compute the gradients of the FFD-GAN parameterization with respect to its design variables (\ie, the latent vector).

Figure~\ref{fig:opt_history_combined} shows the optimization history of three parameterizations with different design variable settings. For the latent dimensions from 5 to 20, The FFD-GAN can find the solution with $C_L/C_D>150$ within the 10 initial evaluations, while FFD and B-spline surface parameterizations take at least 60 evaluations to achieve the same level of $C_L/C_D$. This accelerated convergence demonstrates the high compactness of the FFD-GAN parameterization. Base on the cumulative regret, the FFD-GAN with 5 latent dimensions, the FFD with $4\times 4\times 2$ control points, and the B-spline with $4\times 18$ control points gives the best optimization performance. By further comparing the three parameterizations under those settings, we see that FFD-GAN not only takes much fewer evaluations to discover a high-performance solution, but also finds better solutions in the end. These results indicate that the high representation capacity and compactness of the FFD-GAN can significantly improve the efficiency in shape optimization.


\begin{figure}[hbt!]
\centering
\includegraphics[width=1\textwidth]{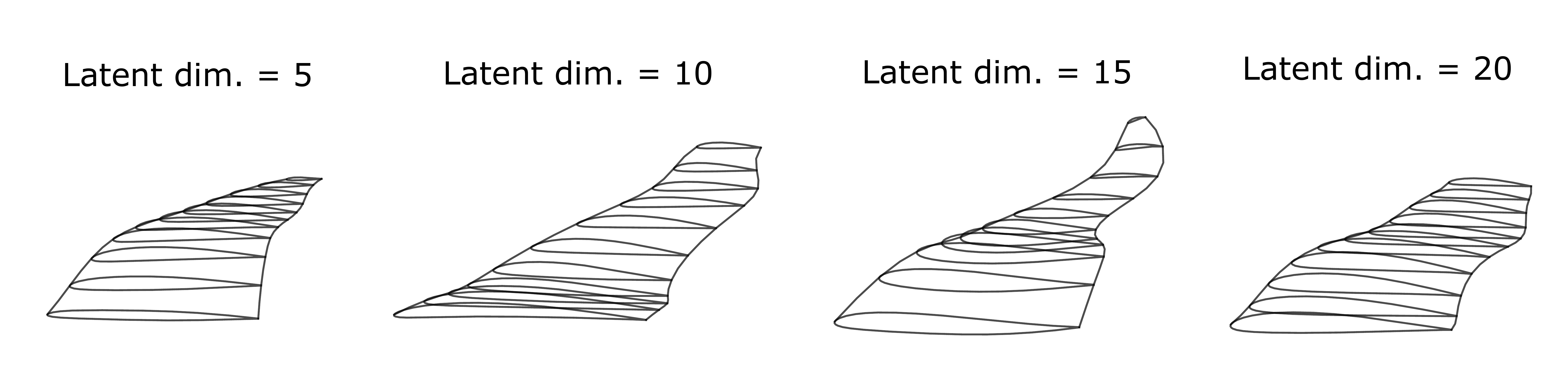}
\caption{Optimized wing designs using FFD-GAN as the parameterization.}
\label{fig:opt_airfoils_ffdgan}
\end{figure}


Figure~\ref{fig:opt_airfoils_ffdgan} shows the geometries of optimized wing designs using FFD-GAN as the parameterization. Note that as a vortex-lattice code, AVL does not predict flow separation~\cite{yang2012aerodynamic} and neglects the influence of viscosity. Thus, the wings shown in Fig.~\ref{fig:opt_airfoils_ffdgan} might not reflect the optimal solution in reality. However, we use AVL here as a black-box simulator to demonstrate the efficiency of different parameterizations. There is no limit on the type of simulators for downstream tasks. In practice, one can use any higher-fidelity CFD simulator to evaluate the performance of aerodynamic shapes.


\section{Conclusions}

We propose a new deep generative model, FFD-GAN, and use it for parameterizing 3D airfoil shapes. This FFD-GAN uses a generator to map a small set of latent variables to 3D shapes represented by surface points. The generator can therefore synthesize 3D geometries from a low-dimensional latent vector. We treat the generator as a parametric model and the latent variables as shape parameters or design variables. To avoid the issue of generating non-smooth surfaces and affect the aerodynamic property, we append an FFD layer to the generator so that it only produces smooth shapes. To demonstrate the efficiency of the FFD-GAN as a new parameterization, we create a wing design dataset using a probabilistic grammar, and perform a series of experiments to test the representation capacity and compactness of FFD-GAN. The results show that, compared to FFD and B-spline surface parameterizations, FFD-GAN exhibits significantly higher representation compactness and comparable representation capacity.

Although we demonstrated FFD-GAN's performance through an aircraft wing design example, the proposed FFD-GAN can also be used for parameterizing other smooth aerodynamic or hydrodynamic shapes, such as turbine blades/vanes, car body, and hulls. Investigating the performance of FFD-GAN on those use cases could be an interesting direction for future work.

In this work, we did not study the explainability and disentanglement of the latent space in the FFD-GAN. Ideally, an explainable and disentangled latent space should have each dimension capturing the variation of a single attribute (\eg, wing twist and dihedral). This helps designers better understand the parameterization and may facilitate design space exploration. In the future, we will explore ways to disentangle the latent space of FFD-GAN~\cite{chen2016infogan,lin2020infogan} and study the effects of latent space disentanglement on design space exploration tasks.

\bibliography{references}

\end{document}